\documentclass[conference]{IEEEtran}
\IEEEoverridecommandlockouts
\usepackage{cite}
\usepackage{amsmath,amssymb,amsfonts}
\usepackage{algorithmic}
\usepackage{graphicx}
\usepackage{textcomp}
\usepackage{xcolor}
\usepackage{mathtools}
\usepackage{authblk}
\usepackage{hyperref}

\usepackage{amsmath}
\usepackage{makecell}
\usepackage{multirow}
\usepackage{booktabs}
\usepackage{bbding}
\usepackage{array}
\usepackage{subcaption}

\def\BibTeX{{\rm B\kern-.05em{\sc i\kern-.025em b}\kern-.08em
    T\kern-.1667em\lower.7ex\hbox{E}\kern-.125emX}}
\begin{document}

\title{BPCLIP: A Bottom-up Image Quality Assessment from Distortion to Semantics Based on CLIP}





\author{Chenyue Song\textsuperscript{1}, Chen Hui\textsuperscript{2}, Wei Zhang\textsuperscript{1}, Haiqi Zhu\textsuperscript{1,\(\dagger\)}\thanks{\raisebox{0.3ex}{\(\dagger\)} Corresponding Author.\\ This work is supported in part by the Central Guidance for Local Science and Technology Development Fund Projects under grant 2024ZYD0266, in part by the Startup Foundation for Introducing Talent of Nanjing University of Information Science and Technology under grant 2025r029.}, Shaohui Liu\textsuperscript{1}, Hong Huang\textsuperscript{3}, Feng Jiang\textsuperscript{1}\\
\textsuperscript{1}Harbin Institute of Technology, China\\
\textsuperscript{2}Nanjing University of Information Science and Technology, China\\
\textsuperscript{3}Sichuan University of Science \& Engineering, China\\
\{cysong, 19B908079\}@stu.hit.edu.cn, chui@nuist.edu.cn, \\
\{haiqizhu, shliu, fjiang\}@hit.edu.cn, huanghong@suse.edu.cn}

\maketitle

\begin{abstract}
Image Quality Assessment (IQA) aims to evaluate the perceptual quality of images based on human subjective perception.    Existing methods generally combine multiscale features to achieve high performance, but most rely on straightforward linear fusion of these features, which may not adequately capture the impact of distortions on semantic content.    To address this, we propose a bottom-up image quality assessment approach based on the Contrastive Language-Image Pre-training (CLIP, a recently proposed model that aligns images and text in a shared feature space), named BPCLIP, which progressively extracts the impact of low-level distortions on high-level semantics.    Specifically, we utilize an encoder to extract multiscale features from the input image and introduce a bottom-up multiscale cross attention module designed to capture the relationships between shallow and deep features.    In addition, by incorporating 40 image quality adjectives across six distinct dimensions, we enable the pre-trained CLIP text encoder to generate representations of the intrinsic quality of the image, thereby strengthening the connection between image quality perception and human language. Our method achieves superior results on most public Full-Reference (FR) and No-Reference (NR) IQA benchmarks, while demonstrating greater robustness.
\end{abstract}


\section{Introduction}
Image Quality Assessment (IQA) aims to estimate perceptual image quality similar to the human visual system
(HVS) and is widely applied in various fields such as image processing, compression and storage. Depending on whether an original reference image is required, IQA methods are typically categorized into Full-Reference (FR) IQA and No-Reference (NR) IQA.

Many mathematical models have been developed to assess the appearance or quality of images. Although these tools are effective in quantifying degradations such as noise and blurriness, their integration with human language is still inadequate. Recently, CLIP-based methods have shown promise in IQA tasks by aligning language and vision\cite{srinath2024learning,wang2023exploring,zhang2023blind}. However, CLIP\cite{radford2021learning} struggles with quality-aware image representations due to its focus on high-level semantics over low-level features\cite{zhang2023blind,liang2023iterative,li2024sglp,li2025fedkd}. Building on this, multiscale feature extraction emerges as a crucial strategy for enhancing performance. Traditional methods like MS-SSIM \cite{wang2003multiscale}, BRISQUE \cite{mittal2012no}, and NIQE \cite{mittal2012making} downsample images to multiple scales and compute quality scores in parallel, but they often fail to derive meaningful quality representations from low-resolution images. In contrast, top-down methods enhance feature extraction by using high-level features to guide low-level features \cite{chen2024topiq} and building feature pyramids from top to bottom. Despite being more effective for multiscale extraction, top-down methods also have similar limitations: 1) they overlook how low-level distortions can disrupt high-level semantic information; 2) two images with different distortions can exhibit similar high-level semantic features, which complicates the process of using these high-level features to guide the extraction of low-level features.
\begin{figure}[t]
	\includegraphics[width=1\linewidth]{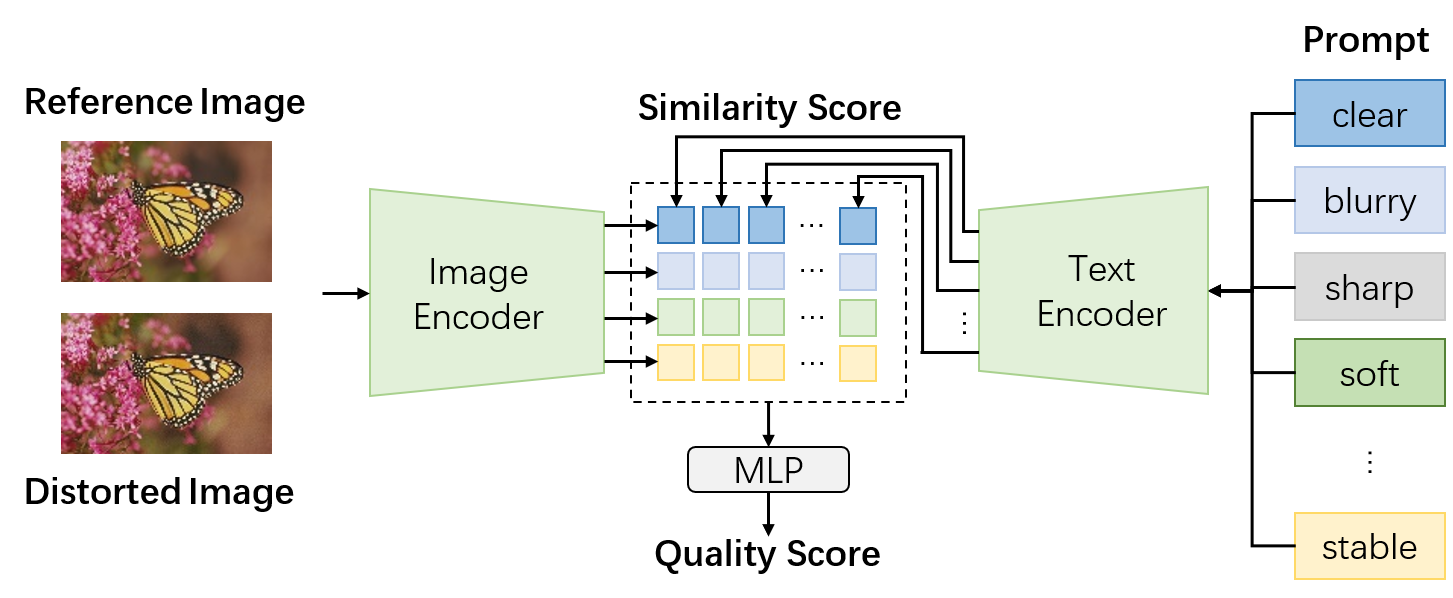}
	\caption{Overview of the final quality score computation strategy. The quality score is computed by applying the softmax function to the similarity between features of 40 image quality adjectives and image features at different hierarchical levels.}
	\label{fig1}
\end{figure}
\begin{figure*}[t]
	\includegraphics[width=1\linewidth]{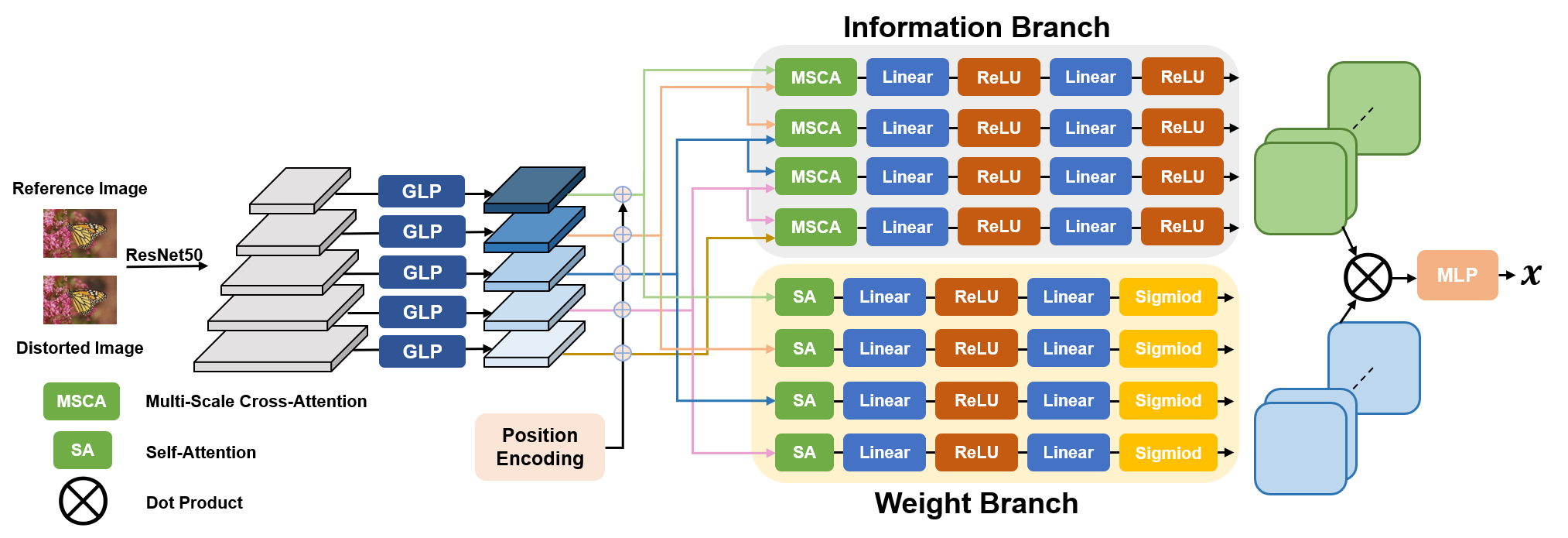}
	\caption{Architecture overview of the proposed BPCLIP Image Encoder. Five distinct levels of features were extracted from the backbone network and progressively propagated through the information branch in a bottom-up manner.}
	\label{fig3}
\end{figure*}

To address this issue, we propose BPCLIP, a bottom-up quality perception method based on CLIP. A key innovation of our method is the multiscale cross attention (MSCA) mechanism, which effectively propagates image information from low to high levels to Extract the effects of distortion on semantic features. In addition, we developed 40 text descriptors related to image quality factors across six distinct dimensions, enabling the pre-trained CLIP text encoder to generate representations of the image's intrinsic quality, thereby strengthening the connection between image quality perception and human language. Extensive experiments on FR and NR IQA datasets demonstrate BPCLIP's superior performance and lower complexity. Our contributions are summarized as follows:
\begin{itemize}
	\item[$\bullet$] We introduce a bottom-up method that leverages multiscale features to progressively capture the impact of low-level distortions on high-level semantics. This not only enhances sensitivity to fine-grained details but also strengthens the model's grasp of high-level semantic structures, leading to more accurate distortion detection and evaluation.
	\item[$\bullet$] We integrate a bottom-up image encoder with a CLIP-based text encoder to link the image quality assessment process with human language descriptions.  By calculating the similarity of quality-related adjectives from multiple dimensions with image features, we improve the naturalness and explainability of the assessment.
	\item[$\bullet$] Our proposed BPCLIP is more efficient than the existing state-of-the-art methods. Utilizing a straightforward ResNet50 \cite{he2016deep} backbone, it achieves competitive performance with reduced computational complexity, striking an optimal balance between performance and efficiency.
\end{itemize}
\section{Proposed Approach}
\subsection{Overview}
The proposed model is illustrated in Figure \ref{fig1}, which consists of a bottom-up image encoder and a text encoder. This network can be applied to both FR and NR tasks. In this section, we focus on the FR framework, as the NR framework is a simplified version. Using the distorted and reference images as inputs, the image encoder progressively propagates multiscale features in a bottom-up approach and ultimately projects them into the CLIP feature space. The detailed architecture is described in Section \ref{img_encoder}. CLIP \cite{radford2021learning} is a multimodal vision-language model trained with a contrastive loss function on a large-scale image-text dataset, designed to semantically align images and their corresponding textual descriptions in a shared embedding space. We use the CLIP text encoder to generate feature representations from 40 image quality descriptors across six distinct dimensions and align them with the output of the image encoder by calculating the cosine similarity between these text feature representations and the image features. The final quality score is derived through regression based on these similarities. During training, the pretrained text encoder stays fixed to ensure consistency in text feature representations.
\begin{figure}[t]
	\includegraphics[width=1\linewidth]{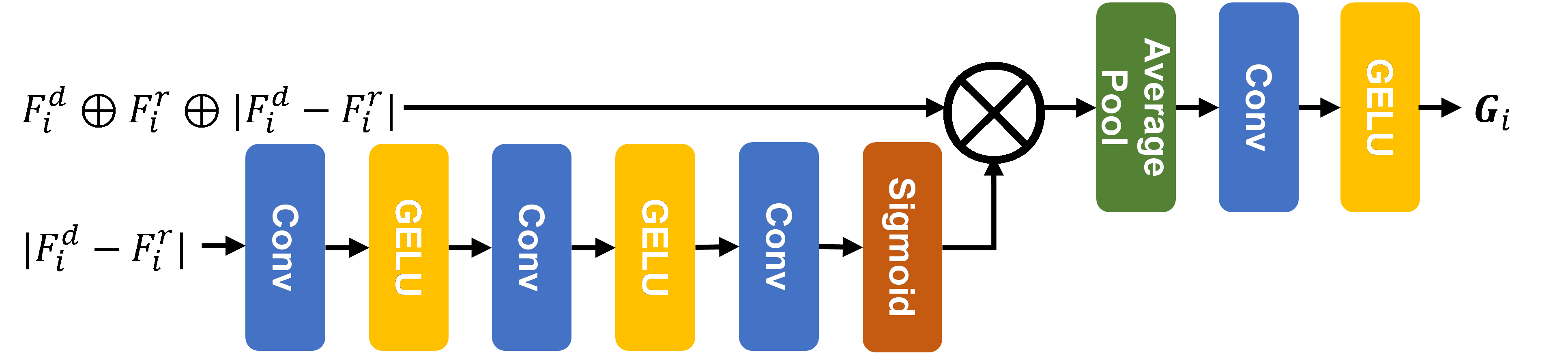}
	\caption{GLP block comprises a mask branch and a feature branch. }
	\label{fig4}
\end{figure}
\begin{table*}[t]
	\centering
	\caption{FR and NR IQA Datasets used for training and evaluation. We resize the SPAQ images so that the shorter side is 448.}
	\begin{center}
		\resizebox{18cm}{!}{
			\begin{tabular}{c c c c c c c c c c}
				\hline
				Type&Dataset&Ref&Dist&Dist Type&Rating&Split&\makecell*[c]{Original Size\\$W×H$}&\makecell*[c]{Resize\\(shorter size)}&\makecell*[c]{Train Size\\(cropped patch)} \\
				\hline
				\multirow{4}{*}{FR}&LIVE&29&779&Synthetic&25k&6:2:2&768×512(typical)&-&384×384	\\
				&CSIQ&30&866&Synthetic&5k&6:2:2&512×512&-&384×384	\\
				&TID2013&25&3k&Synthetic&5k&6:2:2&512×384&-&384×384	\\
				&KADID-10k&81&10.1k&Synthetic&30.4k&6:2:2&512×384&-&384×384	\\
				\hline
				\multirow{3}{*}{NR}&CLIVE&\multirow{3}{*}{-}&1.2k&Authetic&350k&8:2&500×500&-&384×384	\\
				&KonIQ-10k&&10k&Authetic&1.2M&8:2&512×384&-&384×384	\\
				&SPAQ&&11k&Authetic&-&8:2&4k(typical)&448&384×384	\\
				\hline
		\end{tabular}}
		\label{tab1}
	\end{center}
\end{table*}
\subsection{Image Encoder: Bottom-up Network}
\label{img_encoder}
We develop the BPCLIP image encoder network using a bottom-up approach. As shown in Figure \ref{fig3}, the image encoder takes the distorted and reference images as inputs, using ResNet \cite{he2016deep} as the backbone network to extract multiscale features from both the distorted and reference images. We then apply gated local pooling (GLP) \cite{chen2024topiq} to normalize the multiscale features to a consistent spatial size. The information branch uses multiscale cross attention (MSCA) blocks to process features progressively from low-level to high-level, capturing the effects of low-level distortions on high-level semantics. The weight branch incorporates self-attention (SA) blocks to direct the network's focus towards regions with more significant semantic content. Finally, MLP maps the aggregated features to the CLIP feature space.
\subsubsection{Gated Local Pooling}
The input image pair is $(I^d,I^r)\in{\mathbb{R}^{3\times H\times W}}$, and the backbone features for block $i$ are $(F^d_i, F^r_i)\in{\mathbb{R}^{C_i\times H_i\times W_i}}$, where $H_i$ and $W_i$ represent height and width, respectively, and $C_i$ denotes the channel dimension. For ResNet50, $i \in \{1,2,...,n\}$ with $n=5$. Generally, low-level features are roughly twice the size of adjacent high-level features, so we have $H_i=H/2^i$. To enhance simplicity and efficiency, we resize $F_i$ to match the shape of the highest-level feature $F_n$. For the FR task, we define gated convolution as:
\begin{equation}
	F^{mask}_i=\sigma(\phi_i(|F^d_i-F^r_i|))\cdot(F^d_i\oplus F^r_i\oplus |F^d_i-F^r_i|)
\end{equation}
where $\sigma$ is the sigmoid activation function that maps mask values to the range $[0,1]$, $\phi_i$ denotes the bottleneck convolution block, and $\oplus$ indicates the concatenation operation. For a detailed illustration, refer to Figure \ref{fig4}.
For NR tasks, we apply the same gated convolution formula as described below:
\begin{equation}
	F^{mask}_i=\sigma(\phi_i(F_i))\cdot(W_fF_i)
\end{equation}
the masked features $F^{mask}_i$ undergo window average pooling and linear dimensionality reduction to produce the features $G_i \in \mathbb{R}^{D \times H_n \times W_n}$ for the subsequent block, where $D$ represents the dimensionality of the features.
\subsubsection{Attention Modules}
We employ scaled dot-product attention \cite{subakan2021attention} as the core component of our attention module. Given three feature vector sets (query, key, value), the attention mechanism first computes the similarity between the query (Q) and key (K) vectors, and then produces the weighted sum of the values (V), as shown in the following formula:
\begin{equation}
	Attn(Q,K,V)=softmax(\frac{QK^T}{\sqrt{d_k}})V
	\label{func4}
\end{equation}
we apply \eqref{func4} in different ways to support the IQA task. The attention modules are divided into two branches: the information branch and the weight branch. The outputs of these branches are multiplied to produce the weighted features. In addition, in Attention Modules, positional information is crucial as an additional factor for querying features across different scales. Therefore, after GLP, we apply the same learnable positional encoding to all feature maps $G_i$, as illustrated in Figure \ref{fig3}.
\paragraph{Information Branch(MSCA)}
After applying GLP, we obtain a set of features at different scales, denoted as ${G_1, \ldots, G_n} \in \mathbb{R}^{(H_n \times W_n) \times D}$. In this context, the query feature $Q$ in formula (\ref{func4}) inherently guides the output computation. Consequently, our multiscale cross attention mechanism is designed by generating $Q$, $K$, and $V$ from features at various scales, as follows:
\begin{align}
	G^{\prime}_i & =MSCA(G_i,G_{i+1})\nonumber	\\
	~ & =Attn(W_qG_i,W_kG_{i+1},W_vG_{i+1})+G_i
        \label{inf}
\end{align}
where $i \in \{1, ..., n-1\}$. The MSCA block intuitively identifies the impact of lower-level distortions on the semantic integrity of the image, by detecting the correlation between shallow-level distortions in $G_i$ and deep-level semantics in $G_{i+1}$.
\paragraph{Weight Branch(SA)}
We employ self-attention blocks to enhance $G_i$, and then multiply the output of the information branch processed by the MLP with the weight branch to derive the weighted features:
\begin{align}
	G^{\prime\prime}_i & =SA(G_i)=Attn(W_qG_i,W_kG_i,W_vG_i)+G_i	\label{wei}\\
	G^{\prime\prime\prime}_i & =MLP(G^{\prime}_i)\cdot MLP(G^{\prime\prime}_i)
\end{align}
$G^{\prime\prime}_i$ suppresses irrelevant information while emphasizing critical regions that more significantly influence image quality from a human perspective. It is important to note that the last layer of the weight branch uses a Sigmoid activation function to constrain the weight values to $(0,1)$.
\subsubsection{CLIP Feature Space Mapping}
The feature mapping formula is as follows:
\begin{equation}
	x_i=MLP(G^{\prime\prime\prime}_i)
\end{equation}
\begin{figure*}[t]\centering
    \begin{minipage}[t]{0.19\textwidth}
        \centering
        \includegraphics[width=\textwidth]{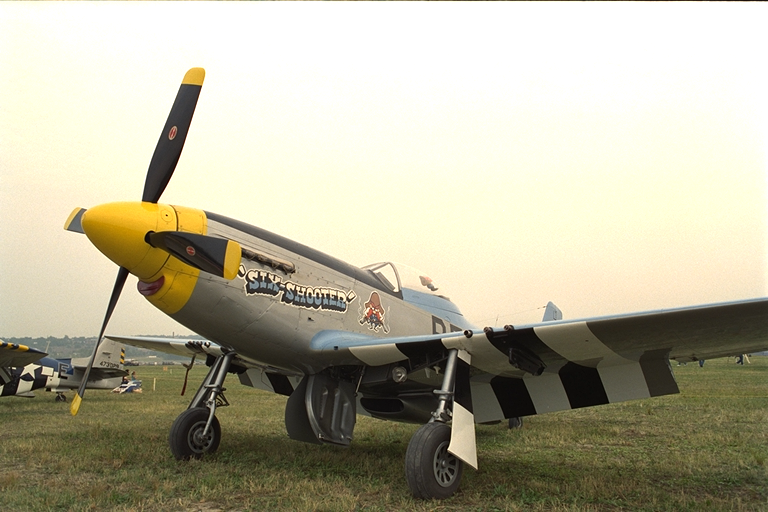}
        \subcaption*{Reference Image}
    \end{minipage}
    \begin{minipage}[t]{0.19\textwidth}
        \centering
        \includegraphics[width=\textwidth]{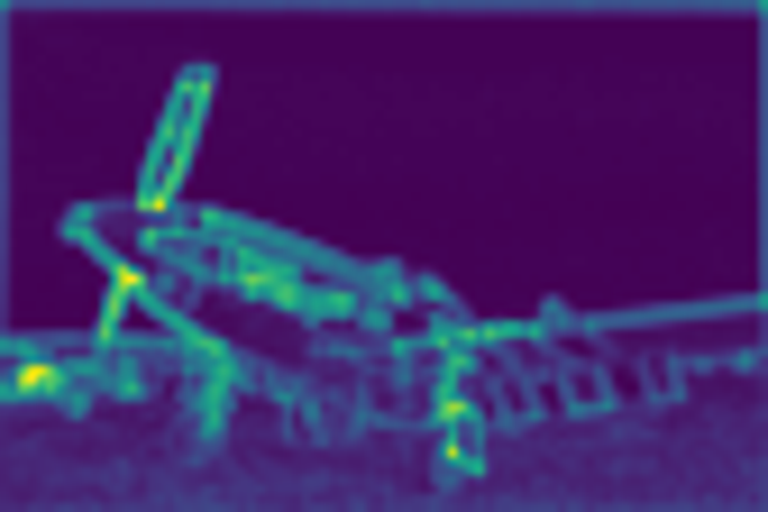}
        \subcaption*{$G^{\prime}_1: MSCA(G_1\rightarrow G_2)$}
    \end{minipage}
    \begin{minipage}[t]{0.19\textwidth}
        \centering
        \includegraphics[width=\textwidth]{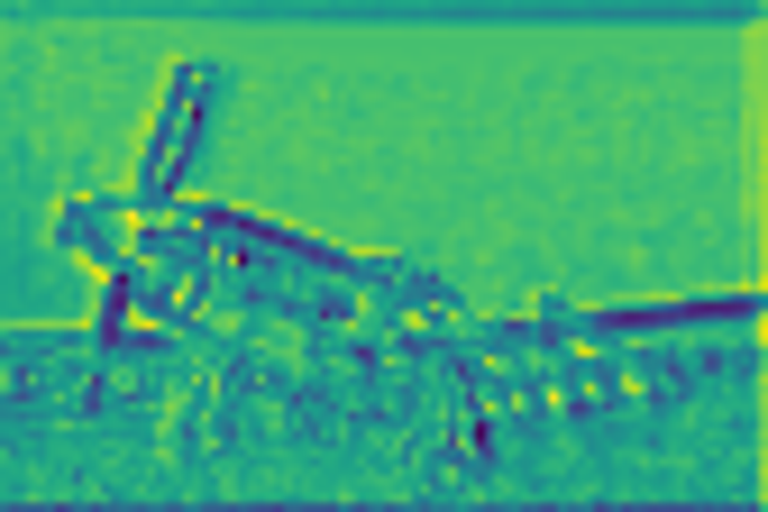}
        \subcaption*{$G^{\prime}_2: MSCA(G_2\rightarrow G_3)$}
    \end{minipage}
    \begin{minipage}[t]{0.19\textwidth}
        \centering
        \includegraphics[width=\textwidth]{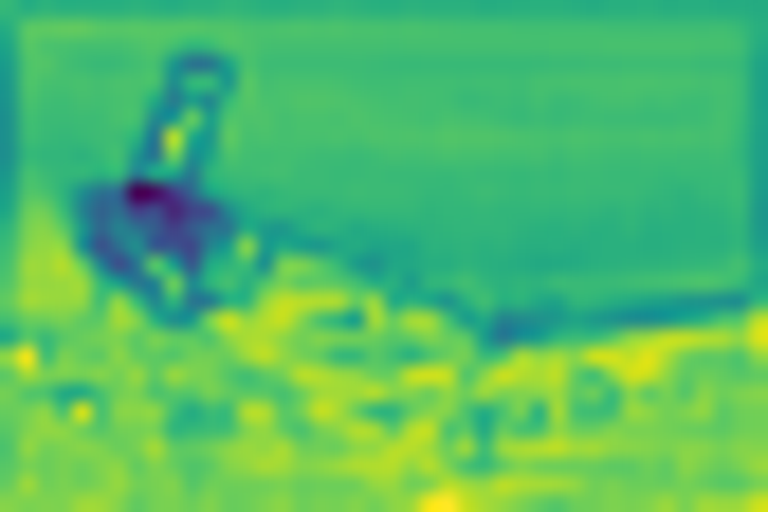}
        \subcaption*{$G^{\prime}_3: MSCA(G_3\rightarrow G_4)$}
    \end{minipage}
    \begin{minipage}[t]{0.19\textwidth}
        \centering
        \includegraphics[width=\textwidth]{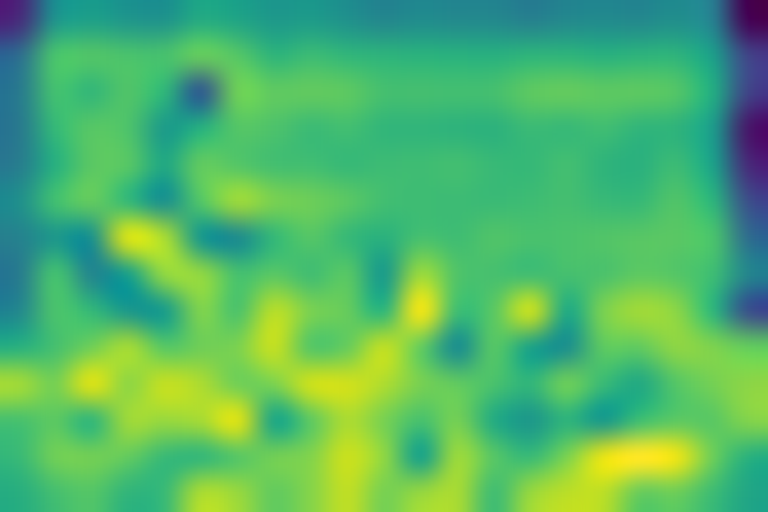}
        \subcaption*{$G^{\prime}_4: MSCA(G_4\rightarrow G_5)$}
    \end{minipage}
    \begin{minipage}[t]{0.19\textwidth}
        \centering
        \includegraphics[width=\textwidth]{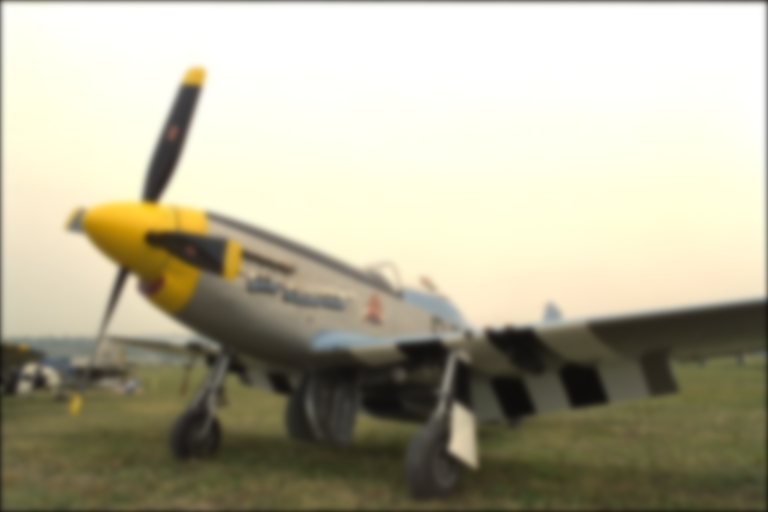}
        \subcaption*{Distorted Image}
    \end{minipage}
    \begin{minipage}[t]{0.19\textwidth}
        \centering
        \includegraphics[width=\textwidth]{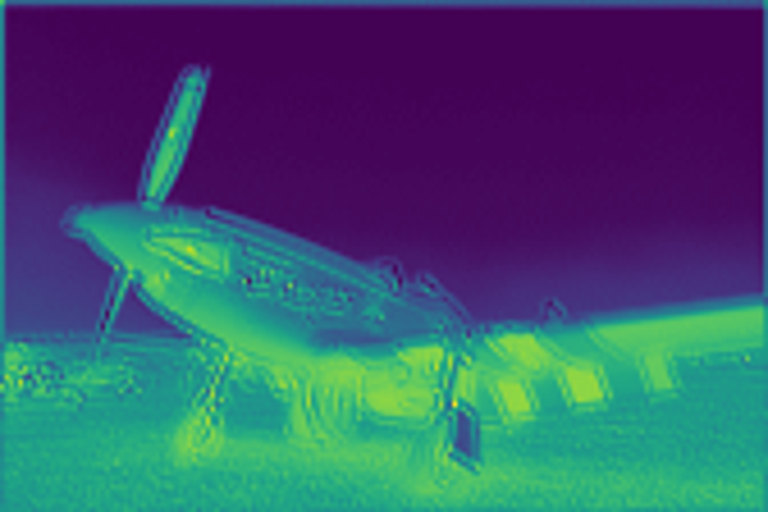}
        \subcaption*{$G^{\prime\prime}_1: SA(G_1)$}
    \end{minipage}
    \begin{minipage}[t]{0.19\textwidth}
        \centering
        \includegraphics[width=\textwidth]{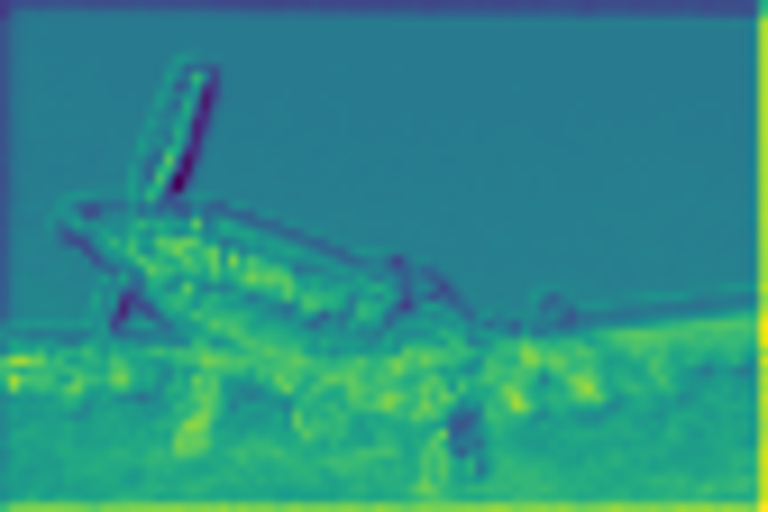}
        \subcaption*{$G^{\prime\prime}_2: SA(G_2)$}
    \end{minipage}
    \begin{minipage}[t]{0.19\textwidth}
        \centering
        \includegraphics[width=\textwidth]{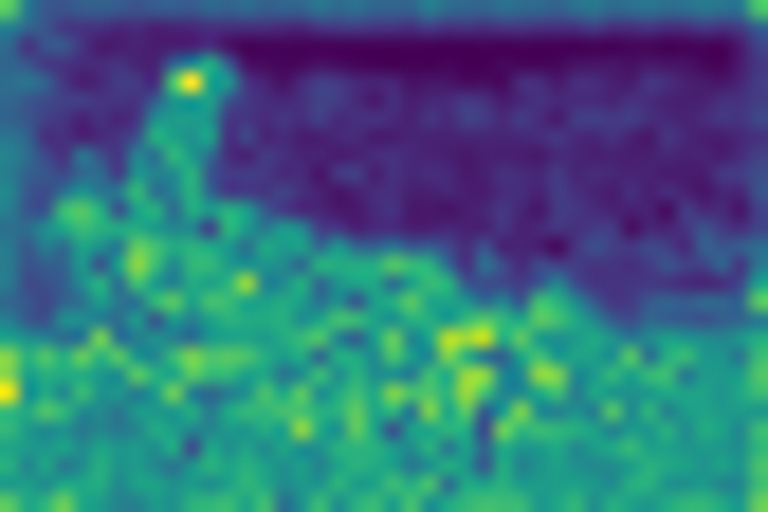}
        \subcaption*{$G^{\prime\prime}_3: SA(G_3)$}
    \end{minipage}
    \begin{minipage}[t]{0.19\textwidth}
        \centering
        \includegraphics[width=\textwidth]{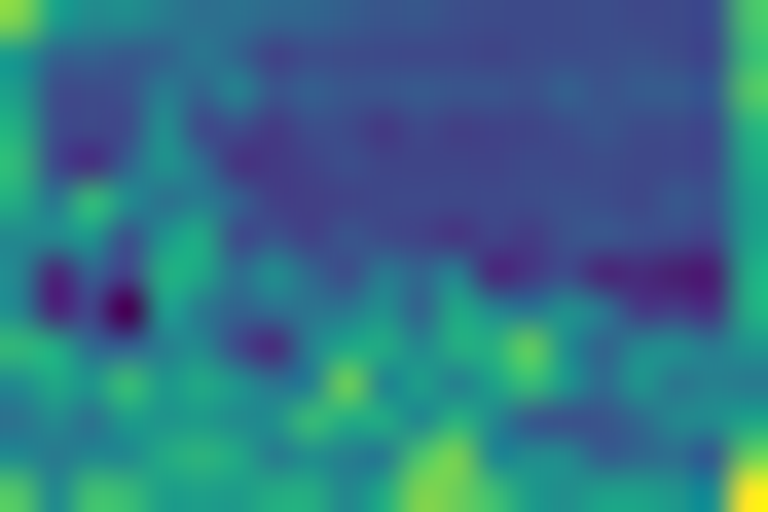}
        \subcaption*{$G^{\prime\prime}_4: SA(G_4)$}
    \end{minipage}
    \caption{Attention-based visualization of Gaussian blur distortion from the LIVE dataset. First row: the output $G^{\prime}_i$ from the information branch in Eq. \ref{inf}; Second row: the output $G^{\prime\prime}_i$ from the weight branch in Eq. \ref{wei}.}
    \label{visual}
\end{figure*}
\subsection{Score Regression}
The final score is computed from the image encoder feature $x$ and text encoder feature $y$:
\begin{align}
	s_i & =\frac{x_i\odot y_i}{\Vert x_i\Vert\cdot \Vert y_i\Vert}	\\
	\hat{p} & =MLP(s_1\oplus s_2\oplus...\oplus s_{n-1})
\end{align}
where $\odot$ denotes the dot product, $\Vert\cdot\Vert$ represents the 2-norm. The cosine similarity $s_i$ between the $i$-th layer feature and the corresponding text prompt is then calculated, and $\hat{p}$ denotes the final score.
\subsection{Loss Function}
We use datasets labeled with Mean Opinion Score (MOS), and we first normalize the MOS to the range $[0,1]$, and then apply the Mean Squared Error (MSE) loss function:
\begin{equation}
	L(\hat{p},p)=\frac{1}{N}\sum_{i=1}^{N}\Vert\hat{p}-p\Vert^2
\end{equation}
\begin{table}[t]
	\caption{Quantitative comparison with related works on public FR benchmarks, including the traditional LIVE, CSIQ, TID2013 and KADID-10k with MOS labels. The best and second results are colored in red and blue, while “-” indicates a non-applicable score.}
	\begin{center}
		\resizebox{8.8cm}{!}{
			\begin{tabular}{c c c c c c c c c}
				\hline
				&\multicolumn{2}{c}{LIVE}&\multicolumn{2}{c}{CSIQ}&\multicolumn{2}{c}{TID2013}&\multicolumn{2}{c}{KADID-10k}	\\
				\cmidrule(lr){2-3}\cmidrule(lr){4-5}\cmidrule(lr){6-7}\cmidrule(lr){8-9}
				Method&PLCC&SRCC&PLCC&SRCC&PLCC&SRCC&PLCC&SRCC	\\
				\hline
				PSNR&0.865&0.873&0.819&0.810&0.677&0.687&0.675&0.676	\\
				SSIM\cite{wang2004image}&0.937&0.948&0.852&0.865&0.777&0.727&0.717&0.724	\\
				MS-SSIM\cite{wang2003multiscale}&0.940&0.951&0.889&0.906&0.830&0.786&0.820&0.826	\\
				VIF\cite{sheikh2006image}&0.960&0.964&0.913&0.911&0.771&0.677&0.687&0.679	\\
				FSIMc\cite{zhang2011fsim}&0.961&0.965&0.919&0.931&0.877&0.851&0.850&0.854	\\
				MAD\cite{larson2010most}&0.968&0.967&0.950&0.947&0.827&0.781&0.799&0.799	\\
				VSI\cite{zhang2014vsi}&0.948&0.952&0.928&0.942&0.900&0.897&0.687&0.679	\\
				\hline
				DeepQA\cite{kim2017deep}&0.982&0.981&0.965&0.961&0.947&0.939&0.891&0.897	\\
				PieAPP\cite{prashnani2018pieapp}&0.979&0.977&0.975&0.973&0.946&0.945&0.836&0.836	\\
				LPIPS\cite{zhang2018unreasonable}&0.934&0.932&0.927&0.903&0.749&0.670&0.839&0.843	\\
				DISTS\cite{ding2020image}&0.980&0.975&0.973&0.965&0.947&0.943&0.886&0.887	\\
				CVRKD\cite{yin2022content}&0.965&0.960&0.965&0.958&0.935&0.928&0.959&0.957	\\
				MANIQA\cite{yang2022maniqa}&0.983&0.982&0.968&0.961&0.943&0.937&0.946&0.944	\\
				JSPL\cite{cao2022incorporating}&0.983&0.980&0.970&0.977&0.949&0.940&\color{blue}0.962&\color{blue}0.960	\\
				AHIQ\cite{lao2022attentions}&\color{red}0.989&\color{blue}0.981&0.976&0.973&\color{blue}0.965&\color{blue}0.961&-&-	\\
				TOPIQ\cite{chen2024topiq}&0.982&\color{blue}0.981&\color{blue}0.979&\color{blue}0.976&0.956&0.953&0.960&0.956	\\
				\hline
				BPCLIP&\color{blue}0.985&\color{red}0.985&\color{red}0.982&\color{red}0.982&\color{red}0.969&\color{red}0.967&\color{red}0.964&\color{red}0.963	\\
				std&$\pm$0.003&$\pm$0.004&$\pm$0.003&$\pm$0.002&$\pm$0.013&$\pm$0.011&0.014&0.013	\\
				\hline
		\end{tabular}}
		\label{tab2}
	\end{center}
\end{table}
\section{Experiments}
\subsection{Implementation Details}
\subsubsection{Datasets and Metrics}
We conduct experiments on several public benchmarks, detailed in Table \ref{tab1}. For FR datasets, we use LIVE\cite{sheikh2006statistical}, CSIQ\cite{larson2010most}, TID2013\cite{ponomarenko2013color}, and KADID-10k\cite{lin2019kadid}. For NR datasets, we use CLIVE\cite{ghadiyaram2015massive}, KonIQ-10k\cite{hosu2020koniq}, and SPAQ\cite{fang2020perceptual}. We resize the SPAQ images so that the shorter side is 448 as shown in Table \ref{tab1}. We perform random splits 10 times and report the mean and standard deviation. For FR datasets, image segmentation is performed based on the reference image to prevent overlapping content. We evaluate performance using Spearman's rank-order correlation coefficient (SRCC) and Pearson's linear correlation coefficient (PLCC), which assess the monotonicity and accuracy of predictions, respectively. Higher values for SRCC and PLCC indicate better performance.
\subsubsection{Training Details}
We utilized ResNet50 pretrained on ImageNet as the backbone, freezing the batch normalization layers while fine-tuning the other parameters. Additionally, the pretrained CLIP text encoder was frozen to maintain consistency in text feature representation. Data augmentation techniques, including random cropping and horizontal/vertical flipping, were applied to preserve the intrinsic quality of the images and enhance the model's generalization ability. The AdamW optimizer was used with a weight decay of $10^{-5}$ across all experiments. The initial learning rate was set to $10^{-4}$ for FR datasets and $3 \times 10^{-5}$ for NR datasets. A cosine annealing scheduler was configured with $T_{max} = 50$, $\eta_{min} = 0$, and $\eta_{max} = lr$, where $T_{max}$ represents the maximum number of iterations, and $\eta_{min}$ and $\eta_{max}$ define the minimum and maximum learning rates, respectively. Training is performed for 200 epochs, with the model implemented in PyTorch and trained on a NVIDIA GeForce RTX 3060 GPU.
\begin{table}[t]
	\caption{Comparison of cross-dataset performance on public benchmarks.}
	\begin{center}
		\resizebox{8.8cm}{!}{
			\begin{tabular}{c c c c c c c}
				\hline
				Train dataset&\multicolumn{6}{c}{KADID-10k}	\\
				\cmidrule(lr){2-7}
				Test dataset&\multicolumn{2}{c}{LIVE}&\multicolumn{2}{c}{CSIQ}&\multicolumn{2}{c}{TID2013}	\\
				\cmidrule(lr){2-3}\cmidrule(lr){4-5}\cmidrule(lr){6-7}
				Method&PLCC&SRCC&PLCC&SRCC&PLCC&SRCC	\\
				\hline
				PieAPP\cite{prashnani2018pieapp}&0.908&0.919&0.877&0.892&0.859&0.876	\\
				LPIPS-VGG\cite{zhang2018unreasonable}&0.934&0.932&0.896&0.876&0.749&0.670	\\
				DISTS\cite{ding2020image}&0.954&0.954&0.928&0.929&0.855&0.830	\\
				AHIQ\cite{lao2022attentions}&0.952&\color{blue}0.970&0.955&0.951&0.899&0.901	\\
				TOPIQ\cite{chen2024topiq}&\color{blue}0.955&0.966&\color{blue}0.962&\color{blue}0.966&\color{blue}0.916&\color{blue}0.915	\\
				\hline
				BPCLIP&\color{red}0.961&\color{red}0.975&\color{red}0.965&\color{red}0.970&\color{red}0.922&\color{red}0.918	\\
				\hline
		\end{tabular}}
		\label{tab3}
	\end{center}
\end{table}
\subsection{Visualization of Multiscale Feature Maps}
In this section, we visualize multiscale attention feature maps to demonstrate how BPCLIP operates in a bottom-up fashion. The BPCLIP image encoder consists of two branches: i) the information branch, which processes features progressively from low-level to high-level, capturing the impact of low-level distortions on higher-level semantics, and ii) the weight branch, which directs the network's attention to regions containing more semantically significant content.

Figure \ref{visual} visualizes the multiscale attention features from $G^{\prime}1$ to $G^{\prime}{4}$ in the information branch and the region importance in multiscale features $G^{\prime\prime}_1, \cdots, G^{\prime\prime}_4$ in the weight branch.

We observe that the MSCA block effectively captures the contours of the airplane at lower levels, while at higher levels, it extracts more abstract features. Additionally, the SA block directs the model's attention to regions with more semantically significant content. This observation indicates that the model progressively extracts distortions in semantic regions in a coarse-to-fine manner, which aligns with human visual perception, and demonstrates that BPCLIP effectively extracts distortion features from critical semantic regions.
\begin{table}[t]
	\caption{Comparison of NR benchmarks:CLIVE, KonIQ-10k and SPAQ.}
	\begin{center}
		\resizebox{8.8cm}{!}{
			\begin{tabular}{c c c c c c c}
				\hline
				&\multicolumn{2}{c}{CLIVE}&\multicolumn{2}{c}{KonIQ-10k}&\multicolumn{2}{c}{SPAQ}	\\
				\cmidrule(lr){2-3}\cmidrule(lr){4-5}\cmidrule(lr){6-7}
				Method&PLCC&SRCC&PLCC&SRCC&PLCC&SRCC	\\
				\hline
				DIIVINE\cite{moorthy2011blind}&0.591&0.588&0.558&0.546&0.600&0.599	\\
				BRISQUE\cite{mittal2012making}&0.629&0.629&0.685&0.681&0.817&0.809	\\
				NIQE\cite{mittal2012making}&0.493&0.451&0.534&0.526&0.712&0.713	\\
				ILNIQE\cite{zhang2015feature}&0.508&0.508&0.537&0.523&0.721&0.713	\\
				\hline
				CONTRIQUE\cite{madhusudana2022image}&0.422&0.394&0.630&0.637&0.680&0.676	\\
				Re-IQA\cite{saha2023re}&0.444&0.418&0.550&0.558&0.618&0.616	\\
				ARNIQA\cite{agnolucci2024arniqa}&0.558&0.484&0.760&0.741&0.797&0.789	\\
				CL-ML\cite{babu2023no}&0.525&0.507&0.645&0.645&0.702&0.701	\\
				MUSIQ\cite{ke2021musiq}&-&-&0.928&0.916&\color{blue}0.918&\color{blue}0.915	\\
				TOPIQ\cite{chen2024topiq}&\color{blue}0.880&\color{blue}0.842&\color{blue}0.935&\color{blue}0.922&0.917&0.914	\\
				\hline
                    GRepQ\cite{srinath2024learning}&0.769&0.740&0.788&0.768&0.809&0.805	\\
                    CLIP-IQA\cite{wang2023exploring}&0.593&0.611&0.733&0.699&0.728&0.733	\\
                    LIQE\cite{zhang2023blind} & 0.749 & 0.770 & 0.806 & 0.807 & 0.813 & 0.816 \\
                    \hline
				BPCLIP&\color{red}0.883&\color{red}0.847&\color{red}0.939&\color{red}0.928&\color{red}0.923&\color{red}0.920	\\
				std&$\pm$0.014&$\pm$0.013&$\pm$0.002&$\pm$0.004&$\pm$0.003&$\pm$0.003	\\
				\hline
		\end{tabular}}
		\label{tab4}
	\end{center}
\end{table}
\begin{table}[t]
	\caption{Results of KonIQ-10k using official split.}
	\begin{center}
			\resizebox{8.8cm}{!}{
					\begin{tabular}{c | c c | c c c | c}
								\hline
								Method&DIIVINE\cite{moorthy2011blind}&BRISQUE\cite{mittal2012no}&KonCept512\cite{hosu2020koniq}&MUSIQ\cite{ke2021musiq}&TOPIQ\cite{chen2024topiq}&BPCLIP	\\
								\hline
								PLCC&0.612&0.707&0.937&0.937&\color{blue}0.939&\color{red}0.945	\\
								SRCC&0.589&0.705&0.912&0.924&\color{blue}0.926&\color{red}0.932	\\
								\hline
							\end{tabular}}
				\label{tab5}
			\end{center}
            \vspace{-3.5mm}
\end{table}
\begin{table}[t]
	\caption{Ablation study for different components in BPCLIP. Experiments are done for both FR and NR datasets.IB is the Information Branch and WB is the Weight Branch.}
	\begin{center}
		\resizebox{8.8cm}{!}{
			\begin{tabular}{c c c c c c c}
				\hline
				\multirow{2}{*}{\makecell[c]{Information Weight \\ Double Branch}}&\multirow{2}{*}{\makecell[c]{Multiscale \\ Cross Attention}}&\multirow{2}{*}{\makecell[c]{CLIP Text \\ Encoder}}&\multicolumn{2}{c}{KADID-10k(FR)}&\multicolumn{2}{c}{KonIQ-10k(NR)}	\\
				\cmidrule(lr){4-5}\cmidrule(lr){6-7}
				&&&PLCC&SRCC&PLCC&SRCC	\\
				\hline
				IB&top-down&\XSolid&0.953&0.950&0.921&0.913	\\
				IB\&WB&top-down&\XSolid&0.955&0.952&0.924&0.914	\\
				IB\&WB&bottom-up&\XSolid&0.960&0.958&0.931&0.919	\\
				IB\&WB&bottom-up&\Checkmark&\color{red}0.964&\color{red}0.963&\color{red}0.939&\color{red}0.928	\\
				\hline
		\end{tabular}}
		\label{tab6}
	\end{center}
\end{table}
\subsection{Comparison with FR Methods}
To validate the superiority of the bottom-up approach, we compare the proposed BPCLIP with various traditional and deep learning methods on the FR benchmark (see Table \ref{tab1}). Our evaluation covers within-dataset experiments, cross-dataset experiments, and computational complexity comparisons.
\subsubsection{Intra-dataset Results of Public Benchmarks}
We performed within-dataset experiments on four benchmarks: LIVE, CSIQ, TID2013 and KADID-10k. Table \ref{tab2} presents the results showing that BPCLIP achieves significant performance on the FR IQA task, demonstrating the effectiveness of the bottom-up feature fusion approach.
\subsubsection{Cross Dataset Experiments}
Furthermore, BPCLIP demonstrates significantly improved generalization with fewer parameters, as shown by the cross dataset experiments presented in Table \ref{tab3}. Comparing the results in Table \ref{tab2} and Table \ref{tab3}, we observe that BPCLIP exhibits greater robustness across various datasets.
\subsubsection{Comparison of computational complexity}
\begin{figure}[t]
        \vspace{-6mm}
	\centering
	\includegraphics[width=\linewidth]{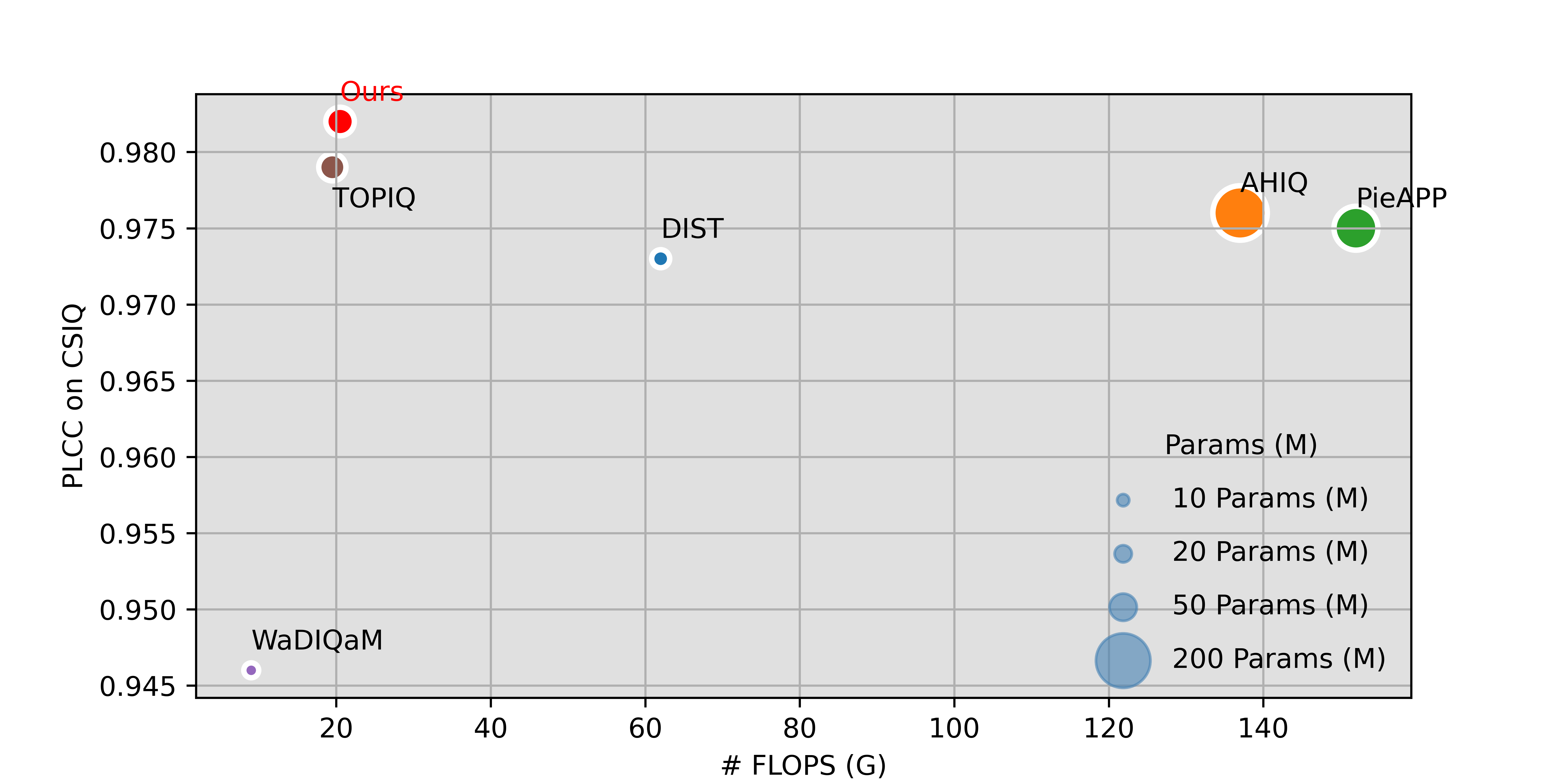}
	\caption{Computational cost (FLOPS) vs. Performance (PLCC) on CSIQ\cite{larson2010most}. The number of parameters is indicated by the circle radius.}
	\label{complex}
\end{figure}
Figure \ref{complex} shows a comparison of computational complexity for recent deep learning-based FR methods. It is evident that BPCLIP achieves the best performance. Moreover, BPCLIP uses ResNet50 as the backbone, and compared to transformer-based methods like AHIQ\cite{lao2022attentions}, its inference time is almost half. Overall, BPCLIP strikes the optimal balance between performance and computational complexity.
\subsection{Comparison with NR Methods}
Due to the absence of reference images, NR-IQA presents greater challenges than FR-IQA.
\subsubsection{Results on NR Benchmarks}The primary NR datasets for real distortions are CLIVE, KonIQ-10k, and SPAQ. The three categories of methods in Table \ref{tab4} are traditional methods, learning-based methods, and CLIP-based methods, with BPCLIP outperforming the other methods, demonstrating that the proposed BPCLIP can effectively handle real distortions even without reference images. In addition, other CLIP-based methods focus solely on optimizing training strategies or loss functions, further confirming the effectiveness of bottom-up feature propagation.
\subsubsection{Results on the KonIQ-10k Official Split}We present the results from 10 random splits of KonIQ10k in Table \ref{tab4}. However, \cite{hosu2020koniq} provides a fixed split in their official code and reports the results. We also present the results obtained using the same settings in Table \ref{tab5}. It is evident that BPCLIP achieves the best performance, further demonstrating the effectiveness and robustness of the proposed BPCLIP.
\subsection{Ablation Study}
Table \ref{tab6} shows that each key component of Bottom: 1) Information Weight Double Branch, 2) Multiscale Cross Attention, and 3) CLIP Text Encoder. We observe that all three components positively impact the results. The MSCA block, in particular, shows the most significant improvement, highlighting the effectiveness of bottom-up feature propagation.
\section{Conclusion}
In this work, we introduce BPCLIP, a bottom-up image quality assessment method based on CLIP. We develop a bottom-up network that progressively captures the influence of distortions on semantic information.The core component of BPCLIP is MSCA, which effectively transfers low-level distortion features to higher semantic layers. In addition, by integrating 40 image quality adjectives across six dimensions, we demonstrate the feasibility of using a pre-trained CLIP text encoder to bridge human language with image quality assessment and improving the model's interpretability. Finally, we conduct extensive experiments on various public benchmarks for both full-reference (FR) and no-reference (NR) scenarios, where BPCLIP consistently achieves state-of-the-art or competitive performance.
\bibliographystyle{IEEEbib}
\bibliography{icme2025references}

\begin{thebibliography}{10}

\bibitem{srinath2024learning}
Suhas Srinath, Shankhanil Mitra, et~al.,
\newblock ``Learning generalizable perceptual representations for data-efficient no-reference image quality assessment,''
\newblock in {\em WACV}, 2024, pp. 22--31.

\bibitem{wang2023exploring}
Jianyi Wang, Kelvin~CK Chan, and Chen~Change Loy,
\newblock ``Exploring clip for assessing the look and feel of images,''
\newblock in {\em AAAI}, 2023, vol.~37, pp. 2555--2563.

\bibitem{zhang2023blind}
Weixia Zhang, Guangtao Zhai, et~al.,
\newblock ``Blind image quality assessment via vision-language correspondence: A multitask learning perspective,''
\newblock in {\em CVPR}, 2023, pp. 14071--14081.

\bibitem{radford2021learning}
Alec Radford, Jong~Wook Kim, et~al.,
\newblock ``Learning transferable visual models from natural language supervision,''
\newblock in {\em ICML}. PMLR, 2021, pp. 8748--8763.

\bibitem{liang2023iterative}
Zhexin Liang, Chongyi Li, et~al.,
\newblock ``Iterative prompt learning for unsupervised backlit image enhancement,''
\newblock in {\em ICCV}, 2023, pp. 8094--8103.

\bibitem{li2024sglp}
Yuqi Li, Yao Lu, et~al.,
\newblock ``Sglp: A similarity guided fast layer partition pruning for compressing large deep models,''
\newblock {\em arXiv preprint arXiv:2410.14720}, 2024.

\bibitem{li2025fedkd}
Yuqi Li, Xingyou Lin, et~al.,
\newblock ``Fedkd-hybrid: Federated hybrid knowledge distillation for lithography hotspot detection,''
\newblock {\em arXiv preprint arXiv:2501.04066}, 2025.

\bibitem{wang2003multiscale}
Zhou Wang, Eero~P Simoncelli, and Alan~C Bovik,
\newblock ``Multiscale structural similarity for image quality assessment,''
\newblock in {\em ACSSC}. Ieee, 2003, vol.~2, pp. 1398--1402.

\bibitem{mittal2012no}
Anish Mittal, Anush~Krishna Moorthy, and Alan~Conrad Bovik,
\newblock ``No-reference image quality assessment in the spatial domain,''
\newblock {\em IEEE TIP}, vol. 21, no. 12, pp. 4695--4708, 2012.

\bibitem{mittal2012making}
Anish Mittal, Rajiv Soundararajan, and Alan~C Bovik,
\newblock ``Making a “completely blind” image quality analyzer,''
\newblock {\em IEEE SPL}, vol. 20, no. 3, pp. 209--212, 2012.

\bibitem{chen2024topiq}
Chaofeng Chen, Jiadi Mo, et~al.,
\newblock ``Topiq: A top-down approach from semantics to distortions for image quality assessment,''
\newblock {\em IEEE TIP}, 2024.

\bibitem{he2016deep}
Kaiming He, Xiangyu Zhang, Shaoqing Ren, and Jian Sun,
\newblock ``Deep residual learning for image recognition,''
\newblock in {\em CVPR}, 2016, pp. 770--778.

\bibitem{subakan2021attention}
Cem Subakan, Mirco Ravanelli, et~al.,
\newblock ``Attention is all you need in speech separation,''
\newblock in {\em ICASSP}. IEEE, 2021, pp. 21--25.

\bibitem{wang2004image}
Zhou Wang, Alan~C Bovik, et~al.,
\newblock ``Image quality assessment: from error visibility to structural similarity,''
\newblock {\em IEEE TIP}, vol. 13, no. 4, pp. 600--612, 2004.

\bibitem{sheikh2006image}
Hamid~R Sheikh and Alan~C Bovik,
\newblock ``Image information and visual quality,''
\newblock {\em IEEE TIP}, vol. 15, no. 2, pp. 430--444, 2006.

\bibitem{zhang2011fsim}
Lin Zhang, Lei Zhang, et~al.,
\newblock ``Fsim: A feature similarity index for image quality assessment,''
\newblock {\em IEEE TIP}, vol. 20, no. 8, pp. 2378--2386, 2011.

\bibitem{larson2010most}
Eric~C Larson and Damon~M Chandler,
\newblock ``Most apparent distortion: full-reference image quality assessment and the role of strategy,''
\newblock {\em JEI}, vol. 19, no. 1, pp. 011006--011006, 2010.

\bibitem{zhang2014vsi}
Lin Zhang, Ying Shen, and Hongyu Li,
\newblock ``Vsi: A visual saliency-induced index for perceptual image quality assessment,''
\newblock {\em IEEE TIP}, vol. 23, no. 10, pp. 4270--4281, 2014.

\bibitem{kim2017deep}
Jongyoo Kim and Sanghoon Lee,
\newblock ``Deep learning of human visual sensitivity in image quality assessment framework,''
\newblock in {\em CVPR}, 2017, pp. 1676--1684.

\bibitem{prashnani2018pieapp}
Ekta Prashnani, Hong Cai, et~al.,
\newblock ``Pieapp: Perceptual image-error assessment through pairwise preference,''
\newblock in {\em CVPR}, 2018, pp. 1808--1817.

\bibitem{zhang2018unreasonable}
Richard Zhang, Phillip Isola, et~al.,
\newblock ``The unreasonable effectiveness of deep features as a perceptual metric,''
\newblock in {\em CVPR}, 2018, pp. 586--595.

\bibitem{ding2020image}
Keyan Ding, Kede Ma, et~al.,
\newblock ``Image quality assessment: Unifying structure and texture similarity,''
\newblock {\em IEEE T-PAMI}, vol. 44, no. 5, pp. 2567--2581, 2020.

\bibitem{yin2022content}
Guanghao Yin, Wei Wang, et~al.,
\newblock ``Content-variant reference image quality assessment via knowledge distillation,''
\newblock in {\em AAAI}, 2022, vol.~36, pp. 3134--3142.

\bibitem{yang2022maniqa}
Sidi Yang, Tianhe Wu, et~al.,
\newblock ``Maniqa: Multi-dimension attention network for no-reference image quality assessment,''
\newblock in {\em CVPR}, 2022, pp. 1191--1200.

\bibitem{cao2022incorporating}
Yue Cao, Zhaolin Wan, et~al.,
\newblock ``Incorporating semi-supervised and positive-unlabeled learning for boosting full reference image quality assessment,''
\newblock in {\em CVPR}, 2022, pp. 5851--5861.

\bibitem{lao2022attentions}
Shanshan Lao, Yuan Gong, et~al.,
\newblock ``Attentions help cnns see better: Attention-based hybrid image quality assessment network,''
\newblock in {\em CVPR}, 2022, pp. 1140--1149.

\bibitem{sheikh2006statistical}
Hamid~R Sheikh, Muhammad~F Sabir, and Alan~C Bovik,
\newblock ``A statistical evaluation of recent full reference image quality assessment algorithms,''
\newblock {\em IEEE TIP}, vol. 15, no. 11, pp. 3440--3451, 2006.

\bibitem{ponomarenko2013color}
Nikolay Ponomarenko, Oleg Ieremeiev, et~al.,
\newblock ``Color image database tid2013: Peculiarities and preliminary results,''
\newblock in {\em EUVIP}. IEEE, 2013, pp. 106--111.

\bibitem{lin2019kadid}
Hanhe Lin, Vlad Hosu, and Dietmar Saupe,
\newblock ``Kadid-10k: A large-scale artificially distorted iqa database,''
\newblock in {\em QoMEX}. IEEE, 2019, pp. 1--3.

\bibitem{ghadiyaram2015massive}
Deepti Ghadiyaram and Alan~C Bovik,
\newblock ``Massive online crowdsourced study of subjective and objective picture quality,''
\newblock {\em IEEE TIP}, vol. 25, no. 1, pp. 372--387, 2015.

\bibitem{hosu2020koniq}
Vlad Hosu, Hanhe Lin, et~al.,
\newblock ``Koniq-10k: An ecologically valid database for deep learning of blind image quality assessment,''
\newblock {\em IEEE TIP}, vol. 29, pp. 4041--4056, 2020.

\bibitem{fang2020perceptual}
Yuming Fang, Hanwei Zhu, et~al.,
\newblock ``Perceptual quality assessment of smartphone photography,''
\newblock in {\em CVPR}, 2020, pp. 3677--3686.

\bibitem{moorthy2011blind}
Anush~Krishna Moorthy and Alan~Conrad Bovik,
\newblock ``Blind image quality assessment: From natural scene statistics to perceptual quality,''
\newblock {\em IEEE TIP}, vol. 20, no. 12, pp. 3350--3364, 2011.

\bibitem{zhang2015feature}
Lin Zhang, Lei Zhang, and Alan~C Bovik,
\newblock ``A feature-enriched completely blind image quality evaluator,''
\newblock {\em IEEE TIP}, vol. 24, no. 8, pp. 2579--2591, 2015.

\bibitem{madhusudana2022image}
Pavan~C Madhusudana, Neil Birkbeck, et~al.,
\newblock ``Image quality assessment using contrastive learning,''
\newblock {\em IEEE TIP}, vol. 31, pp. 4149--4161, 2022.

\bibitem{saha2023re}
Avinab Saha, Sandeep Mishra, and Alan~C Bovik,
\newblock ``Re-iqa: Unsupervised learning for image quality assessment in the wild,''
\newblock in {\em CVPR}, 2023, pp. 5846--5855.

\bibitem{agnolucci2024arniqa}
Lorenzo Agnolucci, Leonardo Galteri, et~al.,
\newblock ``Arniqa: Learning distortion manifold for image quality assessment,''
\newblock in {\em WACV}, 2024, pp. 189--198.

\bibitem{babu2023no}
Nithin~C Babu, Vignesh Kannan, and Rajiv Soundararajan,
\newblock ``No reference opinion unaware quality assessment of authentically distorted images,''
\newblock in {\em WACV}, 2023, pp. 2459--2468.

\bibitem{ke2021musiq}
Junjie Ke, Qifei Wang, et~al.,
\newblock ``Musiq: Multi-scale image quality transformer,''
\newblock in {\em ICCV}, 2021, pp. 5148--5157.

\end{thebibliography}

\end{document}